\documentclass[11pt]{article}
\usepackage[letterpaper,margin=1in]{geometry}
\usepackage[utf8]{inputenc}
\usepackage[T1]{fontenc}
\usepackage{lmodern}
\usepackage{microtype}
\usepackage{amsmath,amssymb,amsthm}
\usepackage{graphicx}
\usepackage{booktabs}
\usepackage{multirow}
\usepackage{enumitem}
\usepackage[round,authoryear]{natbib}
\usepackage[hidelinks]{hyperref}
\usepackage{url}

\title{Rank Portability Does Not Imply Feasibility Portability:\\Target-Specific Evaluation of Joint Hardware Constraints}
\author{Wesley Shu\\The Institute of Energetic Paradigm}
\date{}

\newtheorem{proposition}{Proposition}
\newcommand{\C}{\mathcal{C}}
\newcommand{\risk}{\operatorname{Risk}}
\newcommand{\cov}{\operatorname{Cov}}

\hypersetup{pdfauthor={Wesley Shu},pdftitle={Rank Portability Does Not Imply Feasibility Portability: Target-Specific Evaluation of Joint Hardware Constraints}}

\begin{document}
\maketitle

\begin{abstract}
Cross-device hardware evaluation often relies on a premise that is useful for prediction but unsafe for decision making: if architecture rankings transfer across devices, then a proxy device should support target-side model selection. We stress-test that premise for \emph{joint} latency--energy feasibility. The evaluation spans two public architecture families: 15,625 NAS-Bench-201 architectures with EdgeGPU, Eyeriss, and FPGA resource measurements, and 10,000 common GPT architectures in HW-GPT-Bench over 13 hardware devices. First, on NAS-Bench-201, cross-device latency and energy rank correlations are moderate, while target-comparable feasible-set overlap remains incomplete. A faithful AdaProxy diagnostic substantially improves latency rank correlation (median SRCC $0.623\!\to\!0.859$, $0.721\!\to\!0.929$, and $0.873\!\to\!0.973$), demonstrating that the evaluation is not merely exposing a weak adaptation method. Second, exact finite-sample split-conformal arithmetic reveals an evidence bottleneck hidden by empirical quantile heuristics: a finite one-sided 90\% threshold requires at least nine calibration observations, leaving only one fitting observation under a ten-probe budget. Third, HW-GPT-Bench provides a strong counterexample to equating rank portability with deployment support. Relative to an RTX3080 proxy, target latency SRCC ranges from 0.951 to 0.996 across twelve targets, yet proxy-reuse violation risk ranges from 33.3\% to 100\% under target-comparable joint constraints. With valid 90\% calibration, pooled target-evidence risk remains 15.8\% at 20 probes and 14.8\% at 40 probes, and no common evidence budget through 40 meets the pre-specified target-wise criterion. Together with a previously observed pooled-versus-target authorization reversal on NAS-Bench-201, these results show that rank correlation, pooled frontiers, and target-specific decision support are distinct evaluation objects. We argue that cross-device evaluations should report \emph{target-specific support}: which hardware environments actually support the operating point being claimed.
\end{abstract}

\section{Introduction}
Hardware-aware model selection increasingly depends on transferring measurements across devices. A proxy device can be used to rank candidate architectures, a few target measurements can adapt a predictor, and uncertainty estimates can filter candidates before deployment. These tools are valuable because exhaustive physical measurement is expensive. OneProxy, for example, explicitly exploits cross-device latency monotonicity and adapts a proxy predictor when rank correlation is weak \citep{lu2021oneproxy}. HELP and related few-shot approaches instead measure the actual unseen target and adapt from those observations \citep{lee2021help}. Hardware-aware benchmarks make these questions reproducible across thousands of architectures \citep{li2021hwnasbench,sukthanker2024hwgpt}.

The evaluation problem changes when the output is not a latency estimate but a hard deployment decision. Suppose a system must select the most capable architecture satisfying both latency and energy budgets. The relevant feasible set on target hardware $h$ is
\begin{equation}
  \C_h(B)=\{a:\ell_h(a)\le B_\ell,\;e_h(a)\le B_e\}.
  \label{eq:corridor}
\end{equation}
A candidate outside either coordinate is infeasible. A high rank correlation can still be compatible with substantial disagreement near this two-dimensional boundary, especially when selection chooses an extreme candidate from a large search space. Prior work already shows that hardware-predictor errors can propagate into architecture selection \citep{laube2022predictors}. Our question is more specific: \emph{what evidence actually supports a target-specific feasibility decision?}

This paper is an evaluation-methodology and stress-testing study. It does not propose a new NAS algorithm and it does not claim that cross-device adaptation is ineffective. Instead, it separates three evaluation objects that are often conflated:
\begin{enumerate}[leftmargin=*,itemsep=1pt,topsep=2pt]
\item \textbf{Rank portability:} do proxy and target preserve architecture ordering?
\item \textbf{Feasibility portability:} do they agree on membership in a hard joint resource set?
\item \textbf{Decision support:} does the evidence justify the selected architecture on each target, rather than only on average after pooling targets?
\end{enumerate}

We make four contributions. First, we add explicit rank and feasible-set diagnostics to the NAS-Bench-201 hardware study and show that feasible-set overlap is substantially weaker than rank portability. Second, we reproduce the public OneProxy/AdaProxy adaptation structure in its legitimate latency-ranking domain. Adaptation works: it markedly raises SRCC, which makes the subsequent boundary failures more informative, not less. Third, we replace ``conformal-style'' empirical residual quantiles with the exact finite-sample order-statistic rule and expose the resulting evidence-budget arithmetic. Fourth, we independently replicate the evaluation phenomenon in HW-GPT-Bench, a different architecture family with 10,000 common GPT architectures across 13 devices. There, latency rankings are extraordinarily portable, yet source-only feasibility decisions remain unreliable.

The strongest result is therefore not ``hardware differs.'' It is an evaluation reversal: \emph{evidence can be strong for ranking while weak for authorization}. In the GPT replication, every one of twelve targets has latency SRCC above 0.95 relative to the RTX3080 proxy, yet proxy-reuse target violation risk is at least 33.3\% and reaches 100\%. In the NAS-Bench-201 study, a pooled 40-probe risk of 9.06\% appears to satisfy a 10\% threshold while EdgeGPU and FPGA individually fail it at 12.03\% and 15.00\%. These are different failure modes with the same methodological implication: report target-specific support, not only aggregate performance.

\section{Related work and claim boundary}
\paragraph{Hardware-aware architecture evaluation.}
Hardware-aware NAS incorporates deployment cost directly into model selection \citep{tan2019mnasnet,cai2019proxyless,cai2020ofa}. NAS-Bench-201 and HW-NAS-Bench make architecture and hardware evaluation reproducible in finite search spaces \citep{dong2020nasbench201,li2021hwnasbench}. HW-GPT-Bench extends hardware-aware benchmarking to GPT-family architectures and exposes latency and energy across 13 devices \citep{sukthanker2024hwgpt}. Our contribution is not another benchmark; it is a stress test of how cross-device benchmark evidence is aggregated into deployment claims.

\paragraph{Proxy transfer and target adaptation.}
OneProxy studies cross-device latency rank monotonicity. When monotonicity is high, the searched result on a proxy may transfer; when it is low, AdaProxy adapts the proxy predictor using target observations \citep{lu2021oneproxy}. We therefore use OneProxy as the correct motivation for the rank-portability question, not as a straw-man safety certificate. HELP, Multi-Predict, and recent small-budget search methods use actual target-device evidence and belong on the target-evidence side of the comparison \citep{lee2021help,akhauri2023multipredict,capuano2025budget}.

\paragraph{Uncertainty and selection.}
Conformal prediction provides finite-sample marginal coverage under exchangeability when its order-statistic rule is implemented correctly \citep{vovk2005algorithmic,angelopoulos2023gentle}. Selection complicates interpretation because the candidate of interest is chosen after screening many alternatives; selective prediction and post-selection inference make this distinction explicit \citep{geifman2019selective,jin2024selection}. We use conformal calibration only as an evaluation baseline and do not claim a new conformal theorem.

\section{Evaluation object and protocol}
\subsection{Metrics}
For a decision cell, let $A\in\{0,1\}$ indicate whether the method admits a selected architecture and $U\in\{0,1\}$ indicate whether that admitted architecture violates at least one target resource bound. We report
\begin{equation}
\cov=\mathbb{E}[A],\qquad
\risk=\mathbb{E}[U\mid A=1],
\end{equation}
and, where capability is available, normalized capability regret relative to the best target-feasible architecture. Direct verification has zero lookup-table violation by construction; its empirical cost is therefore coverage and capability opportunity cost, not ``discovered safety.''

\subsection{Target-comparable feasibility regimes}
Absolute latency and energy scales differ dramatically across hardware. To avoid making one target artificially easy, we construct target-specific budgets at oracle feasible-set densities 0.2, 0.4, and 0.6 under balanced, latency-tight, and energy-tight profiles. Each profile rescales the target's median latency and energy and then selects one common scalar threshold so the joint feasible fraction matches the requested density. This preserves the non-compensatory decision in Eq.~\eqref{eq:corridor} while making difficulty comparable.

\subsection{NAS-Bench-201 / HW-NAS-Bench arm}
The first arm uses all 15,625 NAS-Bench-201 architectures with paired latency and energy for EdgeGPU, Eyeriss, and FPGA in HW-NAS-Bench. We evaluate all six ordered source--target pairs. For each pair we report latency SRCC, energy SRCC, a normalized joint-load SRCC, and target-comparable feasible-set Jaccard overlap. We also run a faithful AdaProxy latency diagnostic: Pixel3 is the proxy; the public architecture encoding and Eq.~(3)-style scaling-plus-sparse-residual optimization are used; target train/validation counts follow the public NAS-Bench-201 setup; and the regularization parameter is selected on target-validation SRCC. This diagnostic asks whether the known proxy-adaptation mechanism works before we discuss joint feasibility.

\subsection{Finite-sample target evidence}
For one-sided split conformal with $n$ calibration scores and target miscoverage $\alpha$, we use the $\lceil(n+1)(1-\alpha)\rceil$-th order statistic of the calibration scores augmented with $+\infty$. We never clip an unattainable rank to the largest finite score. Therefore a finite 90\% upper threshold requires at least nine calibration observations. At a total budget of ten measurements, exact 90\% split calibration leaves one fitting observation; at five measurements the finite 90\% threshold is unavailable.

\subsection{Independent HW-GPT-Bench replication}
The second arm uses the official HW-GPT-Bench GPT-small stored ground-truth statistics: sampled latency, sampled energy, and perplexity. The strict common-architecture intersection contains 10,000 architectures. RTX3080 is the proxy and the remaining twelve devices are targets: P100, A100, A40, A6000, H100, RTX2080, V100, three AMD EPYC CPUs, and two Xeon CPUs. We repeat the same 0.2/0.4/0.6 feasible-density regimes and three joint-constraint profiles. Source-only proxy reuse selects the best-perplexity proxy-feasible architecture. Target-evidence mapping uses total budgets 10, 20, and 40, reserves nine observations for exact 90\% calibration, and uses the remaining observations to fit a log resource map from proxy to target. Twenty pre-specified seeds vary the target probes.

\section{Results}
\subsection{Feasible-set portability is weaker than rank portability}
Table~\ref{tab:hwnasrank} reports all ordered NAS-Bench-201 hardware pairs. Latency SRCC ranges from 0.506 to 0.754 and energy SRCC from 0.539 to 0.828, yet median target-comparable feasible-set Jaccard ranges only from 0.483 to 0.626, with minima as low as 0.244. The evaluation object therefore changes before any predictor is fit: preserving global order is easier than preserving membership near a hard joint boundary.

\begin{table}[t]
\caption{Cross-device portability on NAS-Bench-201. Jaccard is computed over target-comparable joint latency--energy feasible sets across the pre-specified density/profile regimes.}
\label{tab:hwnasrank}
\centering
\small
\begin{tabular}{llrrrr}
\toprule
Source & Target & Lat. SRCC & Energy SRCC & Joint SRCC & Median Jaccard\\
\midrule
EdgeGPU & Eyeriss & .506 & .539 & .534 & .483\\
EdgeGPU & FPGA    & .623 & .673 & .652 & .512\\
Eyeriss & EdgeGPU & .506 & .539 & .534 & .483\\
Eyeriss & FPGA    & .754 & .828 & .787 & .626\\
FPGA & EdgeGPU    & .623 & .673 & .652 & .512\\
FPGA & Eyeriss    & .754 & .828 & .787 & .626\\
\bottomrule
\end{tabular}
\end{table}

\subsection{A faithful proxy adaptation improves ranking}
\begin{figure}[t]
\centering
\includegraphics[width=.88\linewidth]{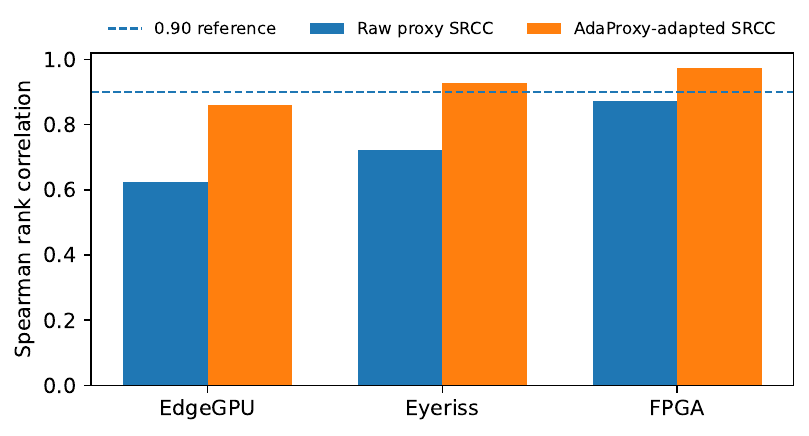}
\caption{Faithful AdaProxy diagnostic. The target-adaptation mechanism substantially improves latency ranking, establishing that the later feasibility failures are not explained by refusing to adapt the proxy.}
\label{fig:ada}
\end{figure}

A negative evaluation of feasibility transfer would be weak if it merely used a poor target-adaptation implementation. Figure~\ref{fig:ada} addresses that concern. Relative to the Pixel3 proxy, faithful AdaProxy adaptation raises median target latency SRCC from 0.623 to 0.859 on EdgeGPU, 0.721 to 0.929 on Eyeriss, and 0.873 to 0.973 on FPGA. All three improve; two exceed 0.90. Thus our claim is not that proxy adaptation fails. Rather, strong rank adaptation is compatible with a separate limitation at the joint feasibility boundary.

\subsection{Finite-sample calibration creates a real evidence bottleneck}
Table~\ref{tab:conf} shows the arithmetic hidden by a small empirical residual quantile. At 90\% nominal coverage, nine calibration points are the minimum for a finite one-sided split-conformal threshold. Consequently, $k=10$ permits only one fitting observation, while $k=5$ cannot produce a finite 90\% threshold at all. At 95\%, the corresponding minimum is 19 calibration observations. This is not an implementation detail: the guarantee itself consumes the measurement budget.

\begin{table}[t]
\caption{Exact finite-sample split-conformal budget arithmetic. ``Fit'' is the maximum remaining target observations after reserving the minimum calibration set.}
\label{tab:conf}
\centering
\small
\begin{tabular}{rrrrr}
\toprule
Nominal coverage & Total $k$ & Min. calibration & Fit & Finite threshold?\\
\midrule
90\% & 5  & 9 & 0  & No\\
90\% & 10 & 9 & 1  & Yes\\
90\% & 20 & 9 & 11 & Yes\\
90\% & 40 & 9 & 31 & Yes\\
95\% & 10 & 19 & 0 & No\\
95\% & 20 & 19 & 1 & Yes\\
\bottomrule
\end{tabular}
\end{table}

\subsection{Independent replication: nearly perfect rankings, unreliable feasibility}
\begin{figure}[t]
\centering
\includegraphics[width=.90\linewidth]{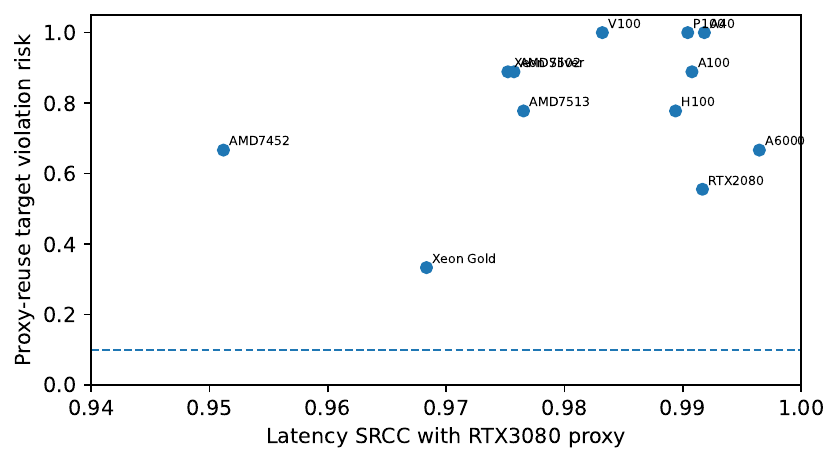}
\caption{Independent HW-GPT-Bench replication. Every target has latency SRCC above 0.95 with the RTX3080 proxy, yet source-only proxy reuse has substantial target violation risk under matched joint latency--energy constraints. Dashed lines show SRCC 0.90 and risk 0.10 only as visual references.}
\label{fig:rankrisk}
\end{figure}

HW-GPT-Bench creates the sharper test. Across twelve targets, latency SRCC with the RTX3080 proxy is between 0.951 and 0.996. Nevertheless, the proxy-feasible best-perplexity selection violates the target's matched joint constraint in 33.3--100\% of evaluation regimes (Figure~\ref{fig:rankrisk}). The contrast is strongest on A6000: latency SRCC is 0.996, yet target violation risk is 66.7\%. A40 has SRCC 0.992 and 100\% violation; P100 has SRCC 0.990 and 100\% violation. High rank portability is therefore not sufficient evidence for hard-boundary portability.

\begin{figure}[t]
\centering
\includegraphics[width=.88\linewidth]{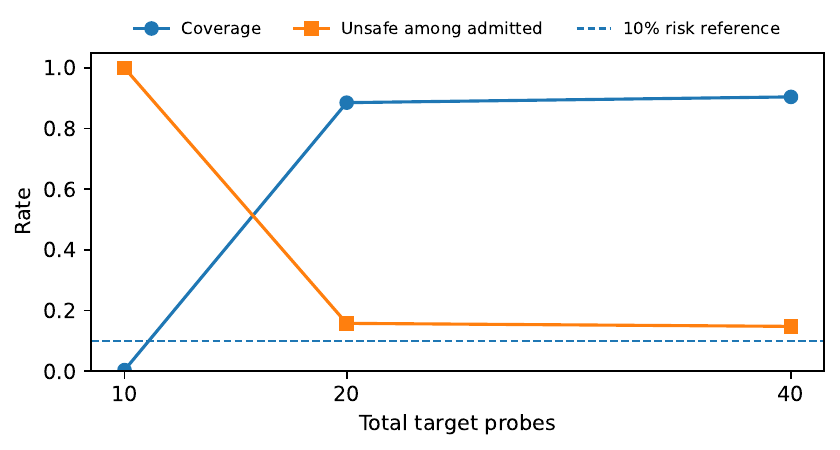}
\caption{HW-GPT-Bench pooled target-evidence frontier using an exact 90\% split-conformal threshold. Coverage recovers with more fitting data, but unsafe-among-admitted risk remains above 10\% through 40 probes. Target-specific support is reported separately in the supplement.}
\label{fig:valid90}
\end{figure}

The target-evidence result is also unfavorable to a universal small-$k$ claim. Under exact 90\% calibration, pooled coverage/risk are 0.37\%/100\% at $k=10$, 88.6\%/15.8\% at $k=20$, and 90.5\%/14.8\% at $k=40$ (Figure~\ref{fig:valid90}). The $k=10$ failure is expected from the finite-sample arithmetic: only one point remains for fitting. More importantly, increasing to 20 or 40 restores coverage but does not reduce pooled conditional violation below 10\%. The pre-specified audit finds no $k\le40$ that simultaneously supports the target-wise criterion on all hardware environments, and it flags a material target-specific support gap of at least 0.20 coverage units.

\subsection{Pooling can change the decision}
The independent GPT replication does \emph{not} produce a pooled-below-threshold reversal; its pooled risk remains above 10\%. The earlier NAS-Bench-201 closure does, which is useful precisely because the two benchmark families expose different failure modes. At $k=40$, a target-ridge uncertainty rule has pooled conditional risk 9.06\%, apparently passing a 10\% criterion, while EdgeGPU and FPGA have target-specific risks 12.03\% and 15.00\% (Eyeriss: 0\%). Pooling would authorize two targets that their own evidence rejects. The new GPT study independently confirms the broader support problem: the pre-specified audit finds a material difference between aggregate and common target support, even though the particular pooled-risk reversal is absent.

\begin{table}[t]
\caption{Cross-benchmark stress-test summary. The two benchmark families expose different manifestations of the same evaluation problem rather than duplicating one artifact.}
\label{tab:replication}
\centering\small
\begin{tabular}{p{1.45cm}p{1.0cm}p{2.2cm}p{2.35cm}p{1.0cm}}
\toprule
Benchmark & Targets & Rank evidence & Decision/support failure & Universal $k\le40$?\\
\midrule
HW-NAS & 3 & SRCC .506--.828; incomplete feasible-set overlap & pooled risk reversal; membership mismatch & No\\
HW-GPT & 12 & latency SRCC .951--.996 & proxy risk .333--1.000; material support gap & No\\
\bottomrule
\end{tabular}
\end{table}

\section{What should hardware evaluations report?}
The experiments support four reporting rules.

\paragraph{Separate ranking from boundary validity.}
Rank metrics such as SRCC answer whether architectures are ordered similarly. They do not answer whether a selected architecture belongs to the target's feasible set. The HW-GPT result makes this distinction unusually clear because correlations above 0.95 coexist with large hard-boundary failure.

\paragraph{Report finite-sample evidence arithmetic.}
A target-probe budget is not fully available for fitting if part of it is reserved for a valid uncertainty statement. For split conformal, the calibration sample required by the desired nominal coverage should be shown explicitly. A heuristic empirical quantile with four calibration points should not be labeled as if it supplied a conventional 90\% finite-sample threshold.

\paragraph{Report target-specific support next to pooled frontiers.}
A pooled risk--coverage curve is descriptive. It becomes decision-relevant only when the target of deployment is sampled from the same mixture and the mixture-level claim is actually the intended object. When the claim is ``this hardware target satisfies risk $\le\epsilon$,'' target-specific support is required. We recommend reporting the range of realizable coverage, the conditional risk at nearby operating points, and whether a common support interval exists across targets.

\paragraph{Treat direct verification as an opportunity-cost baseline.}
If a candidate is measured on the target before admission, zero lookup-table violation is constructional. The question is how much search opportunity was spent to obtain that evidence. In the NAS-Bench-201 closure, active direct verification increased pooled coverage from 73.5\% at five probes to 99.6\% at forty while reducing normalized capability regret from 10.69\% to 2.54\%. This is a useful evidence-sufficiency frontier, not a claim that verification ``discovers'' zero risk.

\section{Limitations}
The two benchmark families differ substantially, which is a strength for replication but prevents a perfectly symmetric experiment. HW-NAS-Bench provides a finite NAS-Bench-201 space with EdgeGPU, Eyeriss, and FPGA latency/energy values; HW-GPT-Bench provides GPT-family ground-truth sampled statistics across many GPU and CPU targets. Both are lookup-table evaluations rather than contemporaneous physical deployment measurements, so they condition on the benchmark's measurement process and do not model thermal drift, firmware changes, repeated-measurement noise, or runtime interference.

The HW-GPT target-evidence baseline is deliberately simple: a log resource map from proxy to target followed by exact split-conformal multiplicative calibration. It is not a claim to outperform HELP, Multi-Predict, or specialized active hardware search. Its purpose is to expose the information accounting under a fixed total probe budget. The faithful AdaProxy experiment is a latency-rank diagnostic, matching the method's legitimate objective; we do not extend its original guarantee to joint latency--energy admission.

The NAS-Bench-201 capability-regret results use aligned CIFAR-100 capability metadata derived through a public development pipeline. The final archival package should retain exact provenance and should not redistribute third-party artifacts whose license is unclear. The new HW-GPT replication uses the official repository's stored ground-truth statistics and its perplexity field. Statistical variability in the closure is over pre-specified probe seeds conditional on the fixed benchmark; it is not population confidence over all possible hardware devices.

\section{Conclusion}
Cross-device hardware evaluation needs a sharper vocabulary. Rank portability, feasibility portability, and target-specific decision support are not interchangeable. A faithful proxy adaptation can substantially improve ranking; exact finite-sample calibration can still consume most of a small evidence budget; and, in a second architecture family, latency rankings above 0.95 can coexist with 33--100\% target violation under hard joint constraints. Pooled summaries add another failure mode: an aggregate point can satisfy a risk threshold while individual targets do not. The practical recommendation is simple: when an evaluation is used to justify deployment, report the \emph{target-specific support} of the operating point being claimed. Aggregate frontiers and proxy correlations remain useful diagnostics, but they should not be treated as authorization evidence by themselves.

\appendix
\section{Claim-scope proposition}
\begin{proposition}[No source-only distribution-free target certificate without a linking assumption]
Fix a source resource map $(\ell_s,e_s)$ and an algorithm that observes only source-side information before admitting an architecture. Without an assumption restricting the relationship between source and target resource maps, there is no nontrivial distribution-free guarantee that every admitted architecture belongs to $\C_t(B)$.
\end{proposition}
\paragraph{Proof.}
Take any source-side observation for which the algorithm admits some architecture $a$. Construct two target worlds that are identical on all source observations. In world 1, set $(\ell_t(a),e_t(a))$ inside $B$; in world 2, change one target coordinate of $a$ to exceed its budget while leaving all source information unchanged. The source-only algorithm makes the same decision in both worlds, so it cannot certify target feasibility in both. A nontrivial target guarantee therefore requires either target evidence or a structural assumption linking source and target resource maps. \hfill$\square$

\section{Executed scientific closure summary}
The pre-specified closure was executed after one solver-only repair: HW-NAS latency was converted from milliseconds to seconds as in the public OneProxy notebook, the unregularized endpoint was solved by least squares, and positive regularization retained the same convex objective with a deterministic solver fallback. No scientific thresholds, seeds, hardware targets, or decision criteria were changed after observing results.

The final verifier confirmed that the faithful AdaProxy diagnostic, finite-sample calibration audit, and 13-device independent replication all executed successfully. Three pre-specified evaluation-failure diagnostics were positive: high HW-GPT rank portability with hard-boundary failure; a material HW-GPT target-support gap; and material NAS-Bench-201 joint feasible-set membership mismatch. The separate HW-GPT pooled-authorization-reversal signal was false; the pooled-versus-target reversal reported in the main text comes from the earlier NAS-Bench-201 closure.

\section{HW-GPT proxy-reuse table}
\begin{table}[h]
\caption{RTX3080 proxy portability on the 12 HW-GPT targets. Risk is the fraction of the nine matched joint-constraint regimes in which the selected proxy-feasible architecture violates the target constraint.}
\centering\small
\begin{tabular}{lrrr}
\toprule
Target & Latency SRCC & Energy SRCC & Target violation risk\\
\midrule
P100 & .9904 & .9363 & 1.000\\
A100 & .9908 & .9502 & .889\\
A40 & .9918 & .9617 & 1.000\\
A6000 & .9965 & .8718 & .667\\
AMD 7452 & .9512 & .9220 & .667\\
AMD 7502 & .9757 & .9543 & .889\\
AMD 7513 & .9765 & .9258 & .778\\
Xeon Gold & .9683 & .8997 & .333\\
Xeon Silver & .9752 & .9524 & .889\\
H100 & .9894 & .9398 & .778\\
RTX2080 & .9917 & .9562 & .556\\
V100 & .9832 & .9771 & 1.000\\
\bottomrule
\end{tabular}
\end{table}

\section{Reproducibility and assets}
The supplementary runner pins the public HW-NAS-Bench blob, downloads the official HW-GPT-Bench GPT-small statistics blob, verifies its Git blob SHA, and freezes the target/proxy roles, target-comparable density regimes, profiles, probe budgets, seeds, and the pre-specified decision rule. HW-NAS-Bench is distributed under the MIT License; HW-GPT-Bench is distributed under Apache-2.0. The supplement should include the executed result CSVs together with the runner before archival submission. The generated row count is not treated as an independent statistical sample size.


\begin{thebibliography}{99}
\bibitem[Tan et~al.(2019)]{tan2019mnasnet} Mingxing Tan et al. MnasNet: Platform-Aware Neural Architecture Search for Mobile. \emph{CVPR}, 2019.
\bibitem[Cai et~al.(2019)]{cai2019proxyless} Han Cai, Ligeng Zhu, and Song Han. ProxylessNAS: Direct Neural Architecture Search on Target Task and Hardware. \emph{ICLR}, 2019.
\bibitem[Cai et~al.(2020)]{cai2020ofa} Han Cai et al. Once-for-All: Train One Network and Specialize it for Efficient Deployment. \emph{ICLR}, 2020.
\bibitem[Dong and Yang(2020)]{dong2020nasbench201} Xuanyi Dong and Yi Yang. NAS-Bench-201: Extending the Scope of Reproducible Neural Architecture Search. \emph{ICLR}, 2020.
\bibitem[Li et~al.(2021)]{li2021hwnasbench} Chaojian Li et al. HW-NAS-Bench: Hardware-Aware Neural Architecture Search Benchmark. \emph{ICLR}, 2021.
\bibitem[Lu et~al.(2021)]{lu2021oneproxy} Bingqian Lu, Jianyi Yang, Weiwen Jiang, Yiyu Shi, and Shaolei Ren. One Proxy Device Is Enough for Hardware-Aware Neural Architecture Search. \emph{Proceedings of the ACM on Measurement and Analysis of Computing Systems}, 5(3), 2021.
\bibitem[Lee et~al.(2021)]{lee2021help} Hayeon Lee et al. HELP: Hardware-Adaptive Efficient Latency Prediction for NAS via Meta-Learning. \emph{NeurIPS}, 2021.
\bibitem[Akhauri and Abdelfattah(2023)]{akhauri2023multipredict} Yash Akhauri and Mohamed S. Abdelfattah. Multi-Predict: Few Shot Predictors For Efficient Neural Architecture Search. 2023.
\bibitem[Capuano et~al.(2025)]{capuano2025budget} Francesco Capuano et al. Searching on a Budget: Hardware-Aware Neural Architecture Search with 10 Latency Probes. 2025.
\bibitem[Laube et~al.(2022)]{laube2022predictors} Kevin Alexander Laube et al. What to Expect of Hardware Metric Predictors in NAS. \emph{AutoML Conference}, 2022.
\bibitem[Sukthanker et~al.(2024)]{sukthanker2024hwgpt} Rhea Sanjay Sukthanker et al. HW-GPT-Bench: Hardware-Aware Architecture Benchmark for Language Models. \emph{NeurIPS Datasets and Benchmarks Track}, 2024.
\bibitem[Vovk et~al.(2005)]{vovk2005algorithmic} Vladimir Vovk, Alex Gammerman, and Glenn Shafer. \emph{Algorithmic Learning in a Random World}. Springer, 2005.
\bibitem[Angelopoulos and Bates(2023)]{angelopoulos2023gentle} Anastasios N. Angelopoulos and Stephen Bates. Conformal Prediction: A Gentle Introduction. \emph{Foundations and Trends in Machine Learning}, 2023.
\bibitem[Geifman and El-Yaniv(2019)]{geifman2019selective} Yonatan Geifman and Ran El-Yaniv. SelectiveNet: A Deep Neural Network with an Integrated Reject Option. \emph{ICML}, 2019.
\bibitem[Jin and Ren(2024)]{jin2024selection} Ying Jin and Zhimei Ren. Confidence on the Focal: Conformal Prediction with Selection-Conditional Coverage. 2024.
\end{thebibliography}
\end{document}